\documentclass[10pt,twocolumn,letterpaper]{article}

\usepackage{cvpr}              

\usepackage{booktabs}
\usepackage{multirow}

\newcommand{\figref}[1]{Fig.~\ref{#1}}
\newcommand{\tabref}[1]{Tab.~\ref{#1}}

\definecolor{cvprblue}{rgb}{0.21,0.49,0.74}
\usepackage[pagebackref,breaklinks,colorlinks,allcolors=cvprblue]{hyperref}

\def\paperID{*****} 
\def\confName{CVPR}
\def\confYear{2026}

\title{Towards Viewpoint-Robust End-to-End Autonomous Driving\\with 3D Foundation Model Priors}

\author{
Hiroki Hashimoto$^{1}$ \quad
Hiromichi Goto$^{2}$ \quad
Hiroyuki Sugai$^{3}$ \\
Hiroshi Kera$^{4,6}$ \quad
Kazuhiko Kawamoto$^{5}$ \\
$^{1,4,5}$Chiba University, $^{2,3}$SUZUCA.AI, $^{6}$National Institute of Informatics\\
{\tt\footnotesize $^{1}$hiroki.hashimoto@chiba-u.jp, $^{2}$goto@suzuca.ai, $^{3}$sugai@suzuca.ai}\\
{\tt\footnotesize $^{4}$kera@chiba-u.jp, $^{5}$kawa@faculty.chiba-u.jp}
}

\begin{document}
\maketitle
\begin{abstract}
Robust trajectory planning under camera viewpoint changes is important for scalable end-to-end autonomous driving.
However, existing models often depend heavily on the camera viewpoints seen during training.
We investigate an augmentation-free approach that leverages geometric priors from a 3D foundation model.
The method injects per-pixel 3D positions derived from depth estimates as positional embeddings and fuses intermediate geometric features through cross-attention.
Experiments on the VR-Drive camera viewpoint perturbation benchmark show reduced performance degradation under most perturbation conditions, with clear improvements under pitch and height perturbations.
Gains under longitudinal translation are smaller, suggesting that more viewpoint-agnostic integration is needed for robustness to camera viewpoint changes.

\end{abstract}
    
\section{Introduction}
\label{sec:intro}

End-to-end autonomous driving jointly optimizes the full pipeline from sensor input to trajectory planning within a single deep neural network.
This approach avoids information loss and error accumulation inherent in conventional modular 
pipelines~\cite{stp3,uniad,vad,sparsedrive,diffusiondrive,law,world4drive,worldrft,vrdrive}.
However, existing end-to-end autonomous driving models depend heavily on the camera viewpoints present in the training data, 
and their trajectory planning accuracy degrades under unseen camera viewpoints~\cite{vrdrive,oncameralidar}.
When autonomous driving systems are deployed across different vehicle platforms, camera viewpoints inevitably vary because of differences in vehicle types and sensor configurations. 
Adapting existing models to such viewpoint changes typically requires additional data collection and retraining for each platform.
Such adaptation is operationally expensive. Improving robustness to unseen camera viewpoints is therefore critical for scalable end-to-end autonomous driving systems.

\begin{figure}[t]
  \centering
  \includegraphics[width=0.9\linewidth]{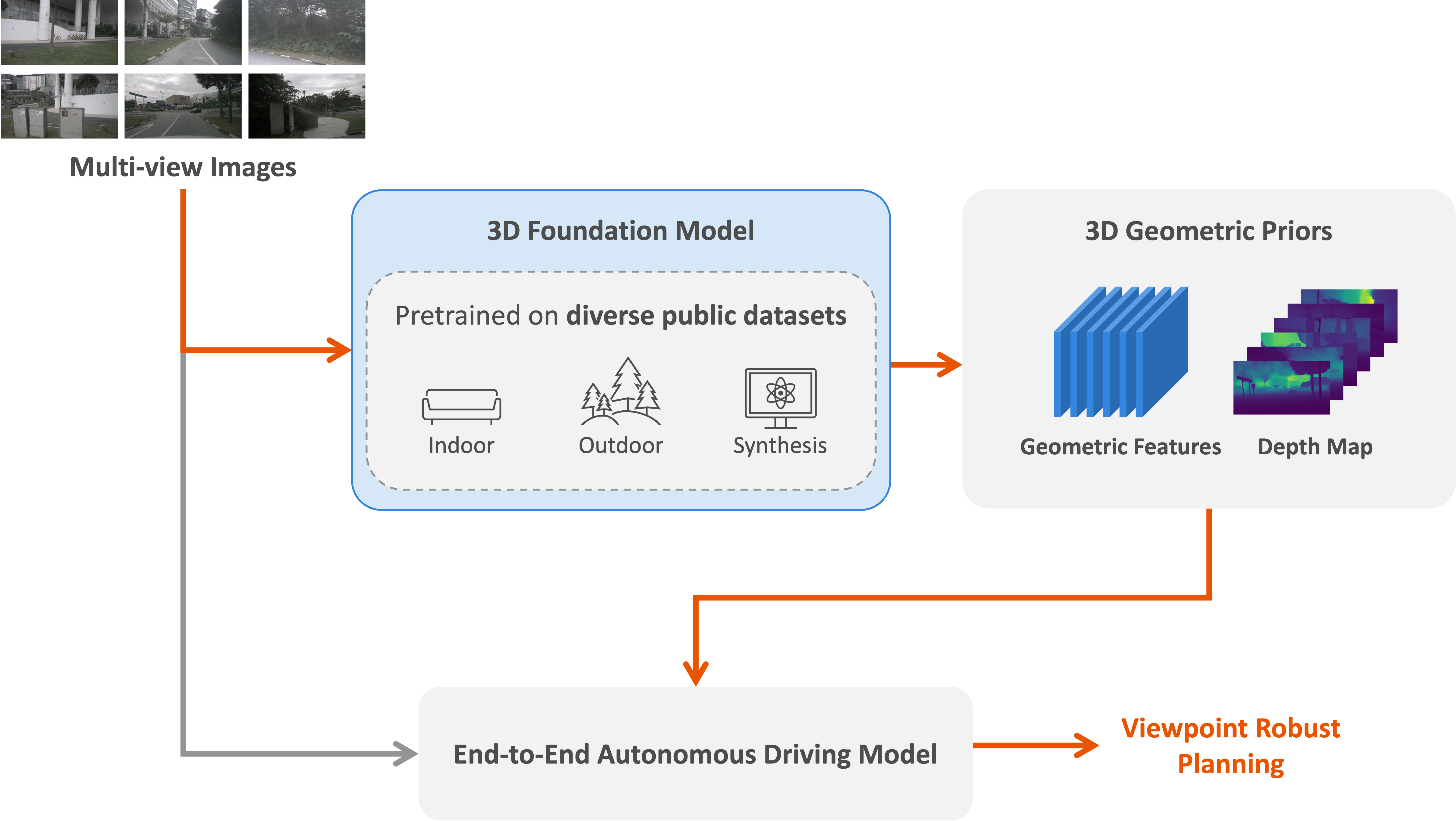}
  \caption{Overview of the proposed method. We extract geometric features and depth estimates from a 3D foundation model and integrate them into an end-to-end autonomous driving model to improve robustness against camera viewpoint changes.}
  \label{fig:teaser}
\end{figure}

To address this challenge, prior work has mainly improved viewpoint robustness by increasing viewpoint diversity during training.
Stelzer et al.~\cite{oncameralidar} show that sensor configuration mismatch degrades end-to-end driving performance, while multi-configuration training improves robustness.
VR-Drive~\cite{vrdrive} improves generalization to camera viewpoint perturbations through
novel view synthesis with 3D Gaussian Splatting (3DGS)~\cite{3dgs}.
However, viewpoint augmentation may not generalize well beyond the camera configurations covered during training.
Augmentation-free approaches are therefore a promising direction for improving robustness to camera viewpoint changes.


In this work, we investigate an augmentation-free approach for improving robustness to camera viewpoint changes in trajectory planning by leveraging geometric priors from a 3D foundation model (\figref{fig:teaser}).
Recent 3D foundation models~\cite{da3,unidepth,metric3d,vggt} acquire generalizable 3D geometric knowledge through large-scale pretraining
and have been applied to downstream driving tasks~\cite{detany3d,world4drive,worldrft}.
To incorporate such geometric knowledge into an end-to-end autonomous driving model,
we introduce two modules (\figref{fig:architecture}).
The first module, 3D Spatial Encoder, computes per-pixel 3D position from DA3~\cite{da3} depth estimates and camera parameters,
and injects these positions as positional embeddings into image features.
The second module, Geometric Feature Fusion, fuses DA3 intermediate features into image features via cross-attention.
These two modules incorporate transferable 3D geometric cues into the driving model.

Evaluation on the VR-Drive viewpoint perturbation benchmark~\cite{vrdrive} shows that the proposed method reduces performance degradation under most perturbation conditions,
with particularly clear gains under pitch and height perturbations.
Performance gains are smaller under longitudinal translation, indicating that some viewpoint changes remain challenging.
These results suggest the importance of a more viewpoint-agnostic integration design for fully leveraging 3D foundation models.


\section{Related Work}
\label{sec:related}

\subsection{Viewpoint Robustness in End-to-End Autonomous Driving}
\label{sec:related_e2e}


Existing end-to-end autonomous driving methods can be broadly grouped into perception-based approaches~\cite{stp3,uniad,vad,sparsedrive,diffusiondrive,vrdrive} and latent world model-based approaches~\cite{law,world4drive,worldrft}. 
Perception-based methods~\cite{stp3,uniad,vad,sparsedrive,diffusiondrive,vrdrive} achieve high planning performance by learning auxiliary perception tasks such as 3D object detection and map construction, but they require large amounts of high-quality 3D annotations. 
Latent world model-based methods~\cite{law,world4drive,worldrft} learn future scene latent representations through self-supervised learning and require less manual annotation, making them suitable for scalable data-driven training.

For robustness to camera viewpoint changes, Stelzer et al.~\cite{oncameralidar} investigate the impact of sensor configurations on end-to-end driving performance in the CARLA simulator~\cite{carla}. 
Their study shows that mismatched sensor configurations between training and testing degrade performance, while multi-configuration training improves robustness.
VR-Drive~\cite{vrdrive} integrates feed-forward 3DGS-based novel view synthesis as an auxiliary task into an end-to-end framework. 
During training, images rendered from randomly sampled camera extrinsics are used together with the original views, increasing viewpoint diversity and improving robustness to camera viewpoint perturbations.
VR-Drive also provides a camera viewpoint perturbation benchmark based on nuScenes~\cite{nuscenes}.

While these approaches improve robustness by augmenting training viewpoints, our work takes an augmentation-free
approach based on geometric priors from a 3D foundation model.

\subsection{3D Foundation Models for Autonomous Driving}
\label{sec:related_foundation}

Recently, 3D foundation models have been developed through large-scale pretraining for 3D geometric tasks such as depth estimation and scene reconstruction~\cite{unidepth,metric3d,vggt,da3}.
Among them, DA3~\cite{da3} uses DINOv2~\cite{dinov2} as its backbone and takes camera intrinsic and extrinsic parameters as camera tokens. This token-based design enables geometrically consistent depth estimation across multiple views and strong generalization to diverse viewpoints and domains.

3D foundation models have also been applied to downstream autonomous driving tasks. 
For object detection, DetAny3D~\cite{detany3d} integrates 
UniDepth-pretrained 2D features with SAM~\cite{sam} features and achieves robust zero-shot monocular 3D object detection, including under unseen camera viewpoints.
For trajectory planning, World4Drive~\cite{world4drive} constructs 3D positional encodings from Metric3D~\cite{metric3d} depth estimates, while WorldRFT~\cite{worldrft} fuses intermediate VGGT~\cite{vggt} features via cross-attention.
These methods improve planning performance, but they do not explicitly address robustness to camera viewpoint changes.
Prior work has not leveraged the geometric knowledge of 3D foundation models to improve such robustness in trajectory planning. 

\section{Method}
\label{sec:method}

\begin{figure*}[t]
  \centering
  \includegraphics[width=0.9\linewidth]{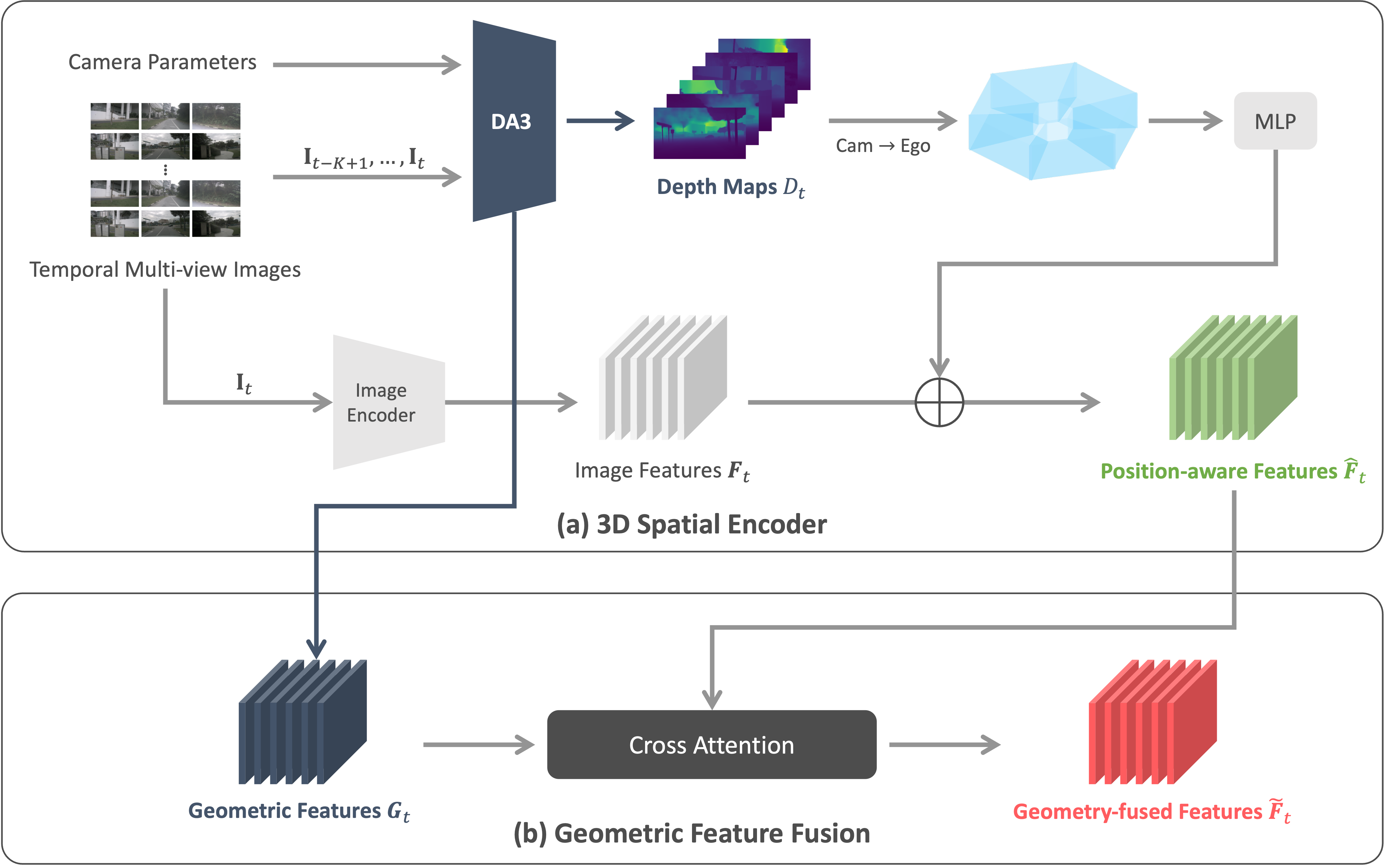}
  \caption{Architecture of the proposed method. (a) 3D Spatial Encoder: computes depth-derived 3D positions from DA3 and camera parameters, and injects them as positional embeddings into image features. (b) Geometric Feature Fusion: fuses DA3 intermediate features into image features via cross-attention.}
  \label{fig:architecture}
\end{figure*}

\subsection{Problem Setting}
\label{sec:problem_setting}

Let $\mathbf{I}_t = \{I^1_t, I^2_t, \ldots, I^N_t\}$ denote the set of $N$ surround-view images captured at time step $t$.
Each camera $i$ has an intrinsic matrix $\mathbf{K}^i \in \mathbb{R}^{3 \times 3}$ and extrinsic parameters $(\mathbf{R}^i \in \mathbb{R}^{3 \times 3},\, \mathbf{d}^i \in \mathbb{R}^{3})$.
We denote the full camera configuration by 
\begin{equation}
\mathcal{C} = \{\mathbf{K}^i, \mathbf{R}^i, \mathbf{d}^i\}_{i=1}^{N}.
\end{equation}
An end-to-end autonomous driving model $F$ takes a temporal sequence of $K$ frames, $\{\mathbf{I}_{t-K+1}, \ldots, \mathbf{I}_t\}$, along with the camera configuration $\mathcal{C}$, and predicts future ego-vehicle waypoints:
\begin{equation}
  \hat{\tau} = F\!\left(\{\mathbf{I}_{t-K+1}, \ldots, \mathbf{I}_t\},\, \mathcal{C}\right),
  \label{eq:model}
\end{equation}
where $\hat{\tau} = \{\hat{\mathbf{p}}_1, \hat{\mathbf{p}}_2, \ldots, \hat{\mathbf{p}}_T\}$ and each $\hat{\mathbf{p}}_j = (x_j, y_j) \in \mathbb{R}^{2}$ denotes the planned ego-vehicle position at future time $j$.

In our setting, training data are collected under a fixed camera configuration $\mathcal{C}_\text{train}$,
whereas test-time deployment involves a different camera configuration $\mathcal{C}_\text{test} \neq \mathcal{C}_\text{train}$.
A change in camera configuration shifts the input distribution from $p(\mathbf{I} \mid \mathcal{C}_{\text{train}})$ to $p(\mathbf{I} \mid \mathcal{C}_{\text{test}})$.
As a result, the predictor $F$ trained under $\mathcal{C}_{\text{train}}$ may produce degraded trajectories $\hat{\tau}$ under $\mathcal{C}_{\text{test}}$.
Our goal is to improve robustness to such viewpoint shifts without requiring additional training data from the test platform.

\subsection{Proposed Method}
\label{sec:method_overview_sub}

We extend World4Drive~\cite{world4drive} for robustness to camera viewpoint changes by integrating geometric priors from DA3~\cite{da3}, without relying on viewpoint augmentation.
As illustrated in \figref{fig:architecture}, the extension consists of two modules.

The first module, \textbf{3D Spatial Encoder}, 
injects depth-derived per-pixel 3D positions as positional embeddings into image features.
The second module, \textbf{Geometric Feature Fusion}, injects DA3 intermediate features into image features via cross-attention.
The first module provides explicit 3D spatial cues, while the second transfers geometric features learned by the pretrained 3D foundation model.
The resulting features are then passed to the planning head for trajectory prediction.

\subsubsection{3D Spatial Encoder}
\label{sec:method_pos3d}
The 3D Spatial Encoder augments image features with depth-derived 3D positional information.
The current surround-view images $\mathbf{I}_t$ are first processed by an image encoder to extract multi-view feature maps $\mathbf{F}_t$.
To estimate scene geometry, DA3 takes the temporal multi-view sequence $\{\mathbf{I}_{t-K+1}, \ldots, \mathbf{I}_t\}$ together with the corresponding camera parameters transformed into the current ego-vehicle coordinate frame. 
DA3 then predicts a depth map $D_t$ for the current frame.

For each pixel $(u, v)$, its 3D position $\mathbf{p}$ in ego-vehicle coordinates is obtained from the depth value $D_t(u,v)$, camera intrinsics $\mathbf{K}$, and extrinsics $(\mathbf{R},\, \mathbf{d})$:
\begin{equation}
  \mathbf{p} = \mathbf{R} \left( \mathbf{K}^{-1} [u,\, v,\, 1]^\top \cdot D_t(u,v) \right) + \mathbf{d}.
  \label{eq:unproject}
\end{equation}
Applying this operation to all pixels yields a 3D point cloud $\mathbf{P}_t$.
The point cloud $\mathbf{P}_t$ is encoded with sinusoidal positional encoding (SPE), and the result is mapped to positional embeddings $\mathbf{E}_t$ via a learnable MLP:
\begin{equation}
  \mathbf{E}_t = \text{MLP}\!\left(\text{SPE}(\mathbf{P}_t)\right).
  \label{eq:pe3d}
\end{equation}
The positional embeddings are then added to the image features:
\begin{equation}
  \hat{\mathbf{F}}_t = \mathbf{F}_t + \mathbf{E}_t.
  \label{eq:feat_aug}
\end{equation}
This operation provides the planner with explicit 3D spatial cues derived from the pretrained depth model.

\subsubsection{Geometric Feature Fusion}
\label{sec:method_da3fusion}

The Geometric Feature Fusion (GFF) module injects DA3 intermediate features into the image features.
From the same temporal multi-view input used in the 3D Spatial Encoder, DA3 intermediate features $\mathbf{G}_t$ are extracted for the current frame.
$\mathbf{G}_t$ encodes geometrically consistent information across views because DA3 internally processes camera tokens.

Following WorldRFT~\cite{worldrft}, the module fuses DA3 features 
into the image features via cross-attention, 
using $\hat{\mathbf{F}}_t$ as the query and $\mathbf{G}_t$ as the key and value:
\begin{equation}
\tilde{\mathbf{F}}_t = \text{CrossAttention}(\hat{\mathbf{F}}_t,\,\mathbf{G}_t).\label{eq:da3fusion}
\end{equation}
The DA3 parameters are frozen during training to preserve the geometric knowledge acquired through large-scale pretraining.
The geometry-fused features $\tilde{\mathbf{F}}_t$ are then passed to the planning head for trajectory prediction.

\section{Experiment}

\subsection{Setup}

\begin{figure}[t]
  \centering
  \includegraphics[width=\linewidth]{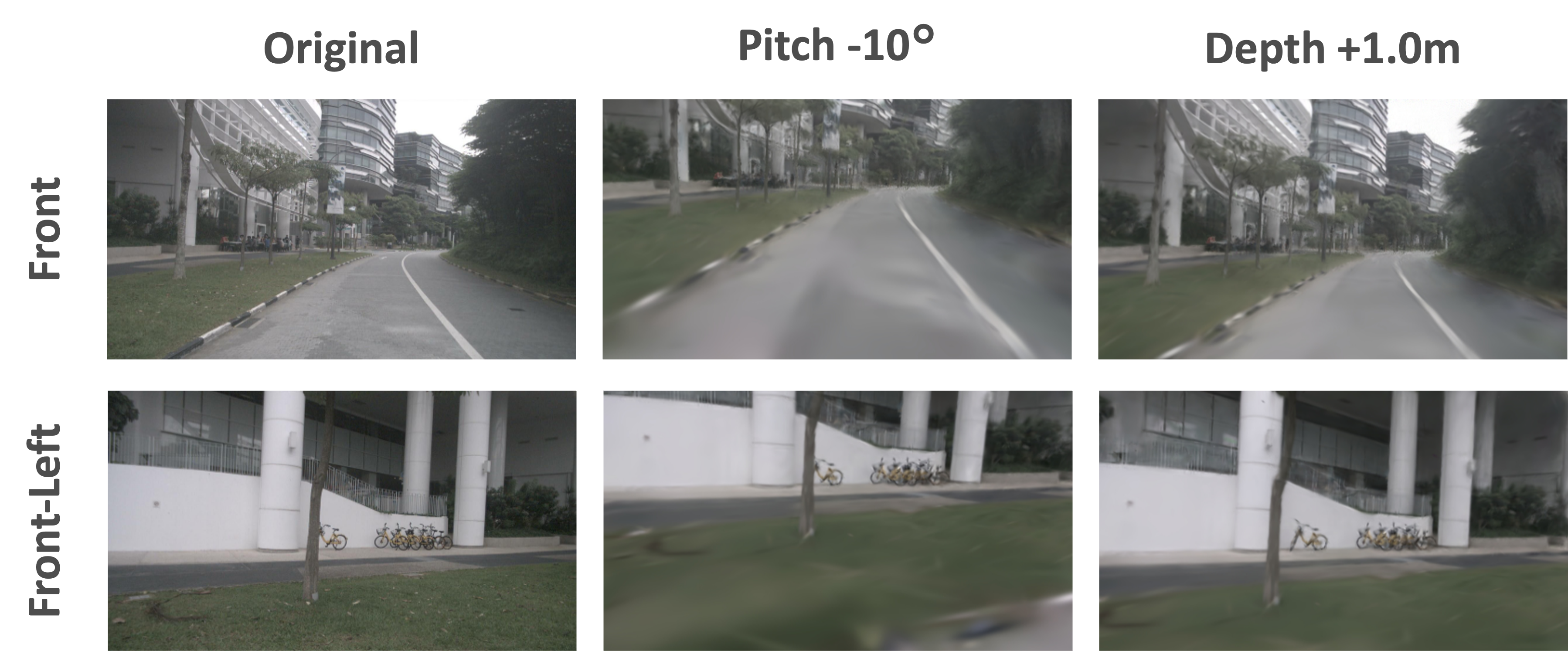}
  \caption{Camera images under different viewpoint perturbation conditions. Each column shows the Original, Pitch $-10^\circ$, and Depth $+1.0$\,m conditions for the front and front-left cameras.}
  \label{fig:camera_sample}
\end{figure}

\paragraph{Dataset.}
We use the VR-Drive~\cite{vrdrive} evaluation framework based on the nuScenes~\cite{nuscenes} dataset to evaluate robustness to viewpoint perturbations.
This framework defines five perturbation conditions: pitch rotation ($-10^\circ$, $+5^\circ$), height translation ($-0.7$\,m, $+1.0$\,m), and depth translation ($+1.0$\,m).
Evaluation images for each perturbation condition are generated via Novel View Synthesis (NVS) using OmniRe~\cite{omnire}. 
Generalization to unseen camera viewpoints is measured
by evaluating models trained on the original nuScenes data using these perturbed images.
\figref{fig:camera_sample} shows example images under representative perturbation conditions.

\paragraph{Metrics.}
Following ST-P3~\cite{stp3}, we adopt L2 displacement error (m) and collision rate (\%) between predicted and ground-truth trajectories.

\paragraph{Baselines.}
We use the reported results of AD-MLP~\cite{admlp}, BEV-Planner~\cite{bevplanner}, VAD~\cite{vad}, SparseDrive~\cite{sparsedrive}, DiffusionDrive~\cite{diffusiondrive}, and VR-Drive~\cite{vrdrive} used in VR-Drive's evaluation.
World4Drive~\cite{world4drive} and the proposed method are trained and evaluated under the same setting.

\paragraph{Training Setup.}
Following World4Drive~\cite{world4drive}, we adopt ResNet-50 as the image encoder with an input image resolution of $360 \times 640$. For DA3~\cite{da3}, we use the Giant-size pretrained model and freeze its parameters during training.
We use AdamW as the optimizer with an initial learning rate of $5 \times 10^{-5}$ and weight decay of $0.01$, with cosine annealing for learning rate scheduling. We train the proposed model for 12 epochs with a batch size of 8, using 8 NVIDIA RTX 6000 Ada GPUs for all experiments.

\subsection{Main Results}

\begin{table*}[t]
\centering
\caption{Comparison of trajectory planning performance under each viewpoint perturbation condition in the VR-Drive evaluation framework.
$*$ denotes methods that use only ego-status without camera images as input.
The Type column indicates P for Perception-based and S for Self-Supervised.}
\label{tab:main_results}
\renewcommand{\arraystretch}{1.15}
\resizebox{\textwidth}{!}{
\setlength\tabcolsep{6.0pt}
\begin{tabular}{l c cc cc cc cc cc cc}
\toprule
\multirow{2}{*}{Method} & \multirow{2}{*}{Type}
& \multicolumn{2}{c}{Original}
& \multicolumn{2}{c}{Pitch $+5^\circ$}
& \multicolumn{2}{c}{Pitch $-10^\circ$}
& \multicolumn{2}{c}{Height $+1.0$\,m}
& \multicolumn{2}{c}{Height $-0.7$\,m}
& \multicolumn{2}{c}{Depth $+1.0$\,m} \\
\cmidrule(lr){3-4} \cmidrule(lr){5-6} \cmidrule(lr){7-8} \cmidrule(lr){9-10} \cmidrule(lr){11-12} \cmidrule(lr){13-14}
& & L2\,[m]$\downarrow$ & Col.\,[\%]$\downarrow$
& L2\,[m]$\downarrow$ & Col.\,[\%]$\downarrow$
& L2\,[m]$\downarrow$ & Col.\,[\%]$\downarrow$
& L2\,[m]$\downarrow$ & Col.\,[\%]$\downarrow$
& L2\,[m]$\downarrow$ & Col.\,[\%]$\downarrow$
& L2\,[m]$\downarrow$ & Col.\,[\%]$\downarrow$ \\
\midrule
AD-MLP$*$
& -
& 0.29 & 0.19
& 0.29 & 0.19
& 0.29 & 0.19
& 0.29 & 0.19
& 0.29 & 0.19
& 0.29 & 0.19 \\
BEV-Planner$*$
& -
& 0.55 & 0.22
& 0.59 & 0.37
& 0.54 & 0.76
& 0.57 & 0.29
& 0.58 & 0.64
& 0.58 & 0.28 \\
VAD
& P
& 0.72 & 0.22
& 0.68 & 0.28
& 1.02 & 0.88
& 0.73 & 0.47
& 0.74 & 0.22
& 0.71 & 0.26 \\
SparseDrive
& P
& 0.61 & 0.08
& 0.66 & 0.15
& 0.96 & 0.23
& 0.87 & 0.54
& 1.01 & 0.30
& 1.27 & 0.31 \\
DiffusionDrive
& P
& 0.57 & 0.08
& 0.67 & 0.11
& 0.96 & 0.24
& 1.46 & 0.81
& 1.21 & 0.20
& 1.57 & 0.41 \\
VR-Drive
& P
& 0.60 & 0.06
& 0.60 & 0.06
& 0.70 & 0.11
& 0.69 & 0.11
& 0.69 & 0.14
& 0.72 & 0.13 \\
\midrule
World4Drive
& S
& 0.48 & 0.26
& 0.83 & 0.48
& 0.64 & 0.36
& 1.06 & 1.29
& 5.01 & 5.44
& 7.10 & 6.40 \\
Ours
& S
& 0.49 & 0.32
& 0.54 & 0.30
& 0.69 & 0.21
& 0.63 & 0.42
& 0.90 & 0.46
& 6.61 & 5.21 \\
\bottomrule
\end{tabular}
}
\end{table*}

\tabref{tab:main_results} shows that the proposed method improves robustness to pitch and height perturbations relative to World4Drive~\cite{world4drive}, while maintaining comparable performance under the original condition. 
In contrast, gains under Depth $+1.0$\,m are smaller, indicating that longitudinal translation remains challenging.
VR-Drive, which augments training viewpoints through novel view synthesis, achieves the best robustness across all perturbation conditions.
These results suggest that augmentation-free geometric priors improve viewpoint robustness, but do not yet match augmentation-based robustness under severe viewpoint shifts.

\begin{figure}[t]
  \centering
  \includegraphics[width=\linewidth]{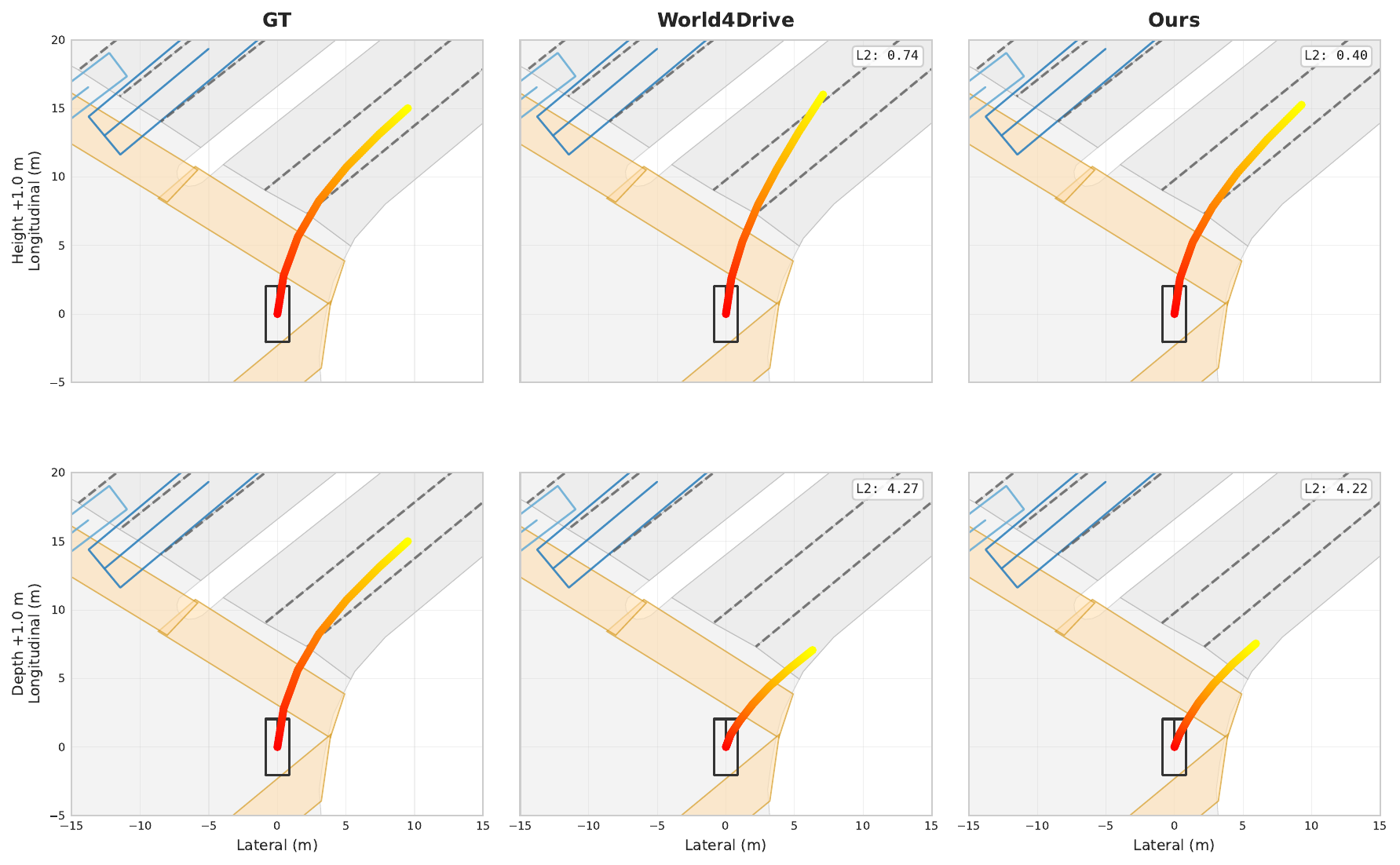}
  \caption{BEV trajectory comparison under viewpoint perturbations. Rows correspond to Height $+1.0$\,m and Depth $+1.0$\,m conditions. Under Height perturbation, the proposed method remains close to the ground truth, whereas World4Drive shows a larger deviation. Under Depth perturbation, both methods fail.}
  \label{fig:bev_trajectory}
\end{figure}

Compared with perception-based end-to-end methods (VAD, SparseDrive, DiffusionDrive), the proposed method does not show a consistent advantage across all perturbation conditions.
\figref{fig:bev_trajectory} shows that the proposed method maintains accurate trajectory prediction under Height perturbation, whereas both World4Drive and our method degrade under Depth perturbation.

\subsection{Further Analysis}

\begin{table*}[t]
\centering
\caption{Ablation of module contributions. The Depth column indicates the depth estimator used in the 3D Spatial Encoder, and the GFF column indicates whether Geometric Feature Fusion is enabled.}

\label{tab:token_only_ablation}
\renewcommand{\arraystretch}{1.15}
\resizebox{\textwidth}{!}{
\setlength\tabcolsep{6.0pt}
\begin{tabular}{lc cc cc cc cc cc cc}
\toprule
\multirow{2}{*}{Depth} & \multirow{2}{*}{GFF}
& \multicolumn{2}{c}{Original}
& \multicolumn{2}{c}{Pitch $+5^\circ$}
& \multicolumn{2}{c}{Pitch $-10^\circ$}
& \multicolumn{2}{c}{Height $+1.0$\,m}
& \multicolumn{2}{c}{Height $-0.7$\,m}
& \multicolumn{2}{c}{Depth $+1.0$\,m} \\
\cmidrule(lr){3-4} \cmidrule(lr){5-6} \cmidrule(lr){7-8} \cmidrule(lr){9-10} \cmidrule(lr){11-12} \cmidrule(lr){13-14}
& & L2\,[m]$\downarrow$ & Col.\,[\%]$\downarrow$
& L2\,[m]$\downarrow$ & Col.\,[\%]$\downarrow$
& L2\,[m]$\downarrow$ & Col.\,[\%]$\downarrow$
& L2\,[m]$\downarrow$ & Col.\,[\%]$\downarrow$
& L2\,[m]$\downarrow$ & Col.\,[\%]$\downarrow$
& L2\,[m]$\downarrow$ & Col.\,[\%]$\downarrow$ \\
\midrule
Metric3D & & 0.48 & 0.26 & 0.83 & 0.48 & 0.64 & 0.36 & 1.06 & 1.29 & 5.01 & 5.44 & 7.10 & 6.40 \\
Metric3D & \checkmark & 0.58 & 0.35 & 0.65 & 0.38 & 0.73 & 0.33 & 0.83 & 0.46 & 0.78 & 0.34 & 4.94 & 5.92 \\
DA3 & & 0.52 & 0.28 & 0.70 & 0.37 & 1.06 & 0.71 & 4.58 & 4.54 & 5.27 & 5.80 & 7.91 & 6.73 \\
DA3 & \checkmark & 0.49 & 0.32 & 0.54 & 0.30 & 0.69 & 0.21 & 0.63 & 0.42 & 0.90 & 0.46 & 6.61 & 5.21 \\
\bottomrule
\end{tabular}
}
\end{table*}

\begin{table}[t]
\centering
\caption{Counterfactual diagnosis under Depth $+1.0$\,m. Camera extrinsic parameters used for 3D positional embeddings are replaced with their training-time values at test time.}

\label{tab:counterfactual}
\begin{tabular}{lcc}
\toprule
Replaced cameras & L2\,[m]$\downarrow$ & Col.\,[\%]$\downarrow$ \\
\midrule
None (Depth $+1.0$\,m) & 6.61 & 5.21 \\
\midrule
Front/rear only & 5.62 & 4.81 \\
Four side cameras & 1.30 & 0.30 \\
All cameras & 0.51 & 0.29 \\
\bottomrule
\end{tabular}
\end{table}

The main results show that the proposed method is effective under pitch and height perturbations, whereas gains under depth perturbation are smaller.
This section investigates the source of this discrepancy through an ablation of the two proposed modules and a counterfactual analysis of the 3D positional embeddings.

\subsubsection{Ablation of Module Contributions}

The proposed method introduces two changes to World4Drive: replacing the depth estimator in the 3D Spatial Encoder with DA3~\cite{da3}, and adding Geometric Feature Fusion (GFF).
To isolate the contribution of each change, we compare the full model
with a configuration that retains Metric3D~\cite{metric3d} as the depth estimator and introduces only GFF (\tabref{tab:token_only_ablation}).

The GFF-only configuration is worse than the baseline under the original condition, but reduces performance degradation under pitch and height perturbations.
This result indicates that GFF is the primary source of the improvement in viewpoint generalization.
By contrast, replacing the depth estimator with DA3 alone degrades performance, while adding GFF mitigates this degradation and recovers gains under several perturbation conditions.
These results show that changing the depth estimator alone does not improve viewpoint generalization.

\subsubsection{Analysis of Distribution Shift in 3D Positional Embedding}

We analyze the limited gains under depth perturbation through a counterfactual diagnosis of the 3D positional embeddings.
The 3D positional embeddings directly depend on camera extrinsic parameters (Eq.~\ref{eq:unproject}).
When camera viewpoints change at test time, the 3D Spatial Encoder produces 3D position distributions that differ from those observed during training.
Training under a single camera configuration can therefore make the model sensitive to the 3D position patterns seen during training.

To verify this hypothesis, we evaluate Depth $+1.0$\,m perturbed images while replacing only the camera extrinsic parameters used for 3D positional embedding computation with their training-time values (\tabref{tab:counterfactual}).
Replacing the four side cameras (front-left, front-right, back-left, back-right) improves L2, and replacing all cameras yields performance close to the original condition.
This result demonstrates that the distributional shift in 3D positional embeddings caused by camera extrinsic changes is the main source of the performance degradation.

\section{Conclusion}
We propose an augmentation-free method for improving robustness to camera viewpoint changes in end-to-end autonomous driving by integrating
geometric priors from a 3D foundation model.
The proposed method uses two modules: the 3D Spatial Encoder for depth-derived 3D positional embeddings and Geometric Feature Fusion for injecting DA3 intermediate features into image features.
Experiments on the VR-Drive viewpoint perturbation benchmark show clear gains under pitch and height perturbations, while gains under longitudinal translation remain limited.

The analysis shows that the main bottleneck comes from 3D positional embeddings that depend explicitly on camera extrinsic parameters.
This result suggests that geometric priors from 3D foundation models are useful for viewpoint robustness, but their effectiveness depends critically on how they are integrated into the driving model.

A promising direction for future work is therefore to construct explicit 3D intermediate representations, such as BEV, voxel, or Gaussian-based scene representations, from the geometric knowledge of 3D foundation models. 
Such representations inherently decouple scene understanding from camera extrinsic parameters and can provide a more viewpoint-agnostic integration to exploit 3D foundation priors.

{
    \small
    \bibliographystyle{ieeenat_fullname}
    \bibliography{main}
}


\end{document}